\documentclass[fleqn,10pt]{SelfArx} 

\usepackage[english]{babel} 

\usepackage{lipsum} 

\usepackage{afterpage}

\usepackage{amsmath}

\usepackage{graphicx}

\makeatletter
\newcommand{\subalign}[1]{%
  \vcenter{%
    \Let@ \restore@math@cr \default@tag
    \baselineskip\fontdimen10 \scriptfont\tw@
    \advance\baselineskip\fontdimen12 \scriptfont\tw@
    \lineskip\thr@@\fontdimen8 \scriptfont\thr@@
    \lineskiplimit\lineskip
    \ialign{\hfil$\m@th\scriptstyle##$&$\m@th\scriptstyle{}##$\hfil\crcr
      #1\crcr
    }%
  }%
}
\makeatother

\definecolor{color1}{RGB}{0,0,90} 
\definecolor{color2}{RGB}{0,20,20} 

\usepackage{hyperref} 
\hypersetup{hidelinks, colorlinks=true, breaklinks=true, urlcolor=blue, citecolor=blue, linkcolor=blue, bookmarksopen=false, pdftitle={Title},pdfauthor={Author}}

\JournalInfo{Rev 9 - 11/18/2021} 
\Archive{First Rev 10/15/2021} 

\PaperTitle{Local Multi-Head Channel Self-Attention for Facial Expression Recognition} 

\Authors{Roberto Pecoraro\textsuperscript{1}, Valerio Basile\textsuperscript{2}, Viviana Bono\textsuperscript{3}, Sara Gallo\textsuperscript{4}} 
\affiliation{\textsuperscript{1}\textit{Department of Computer Science, University of Turin, Italy 10124} - \texttt{roberto.pecoraro@unito.it}; \texttt{robertopecoraro@live.com}} 
\affiliation{\textsuperscript{2}\textit{Department of Computer Science, University of Turin, Italy 10124} - \texttt{valerio.basile@unito.it}} 
\affiliation{\textsuperscript{3}\textit{Department of Computer Science, University of Turin, Italy 10124} - \texttt{bono@di.unito.it}} 
\affiliation{\textsuperscript{4}\textit{Department of Computer Science, University of Turin, Italy 10124} - \texttt{sara.gallo@unito.it}} 

\Keywords{} 
\newcommand{\keywordname}{Keywords} 

\Abstract{Since the Transformer architecture was introduced in 2017 there has been many attempts to bring the self-attention paradigm in the field of computer vision. In this paper we propose a novel self-attention module that can be easily integrated in virtually every convolutional neural network and that is specifically designed for computer vision, the \textit{LHC}: Local (multi) Head Channel (self-attention). \textit{LHC} is based on two main ideas: first, we think that in computer vision the best way to leverage the self-attention paradigm is the channel-wise application instead of the more explored spatial attention and that convolution will not be replaced by attention modules like recurrent networks were in NLP; second, a local approach has the potential to better overcome the limitations of convolution than global attention. With \textit{LHC-Net} we managed to achieve a new state of the art in the famous FER2013 dataset with a significantly lower complexity and impact on the ``host'' architecture in terms of computational cost when compared with the previous SOTA.}

\begin{document}

\flushbottom 

\maketitle 

\tableofcontents 

\thispagestyle{empty} 


\section*{Introduction} 

\addcontentsline{toc}{section}{Introduction} 

The aim of this work is to explore the capabilities of the self-attention paradigm in the context of computer vision, more in particular, in the facial expression recognition. In order to do that we designed a new channel self-attention module, the \textit{LHC}, which is thought as a processing block to be integrated into a pre-existing convolutional architecture. 

It inherits the basic skeleton of the self-attention module from the very well known \textit{Transformer} architecture by Vaswani et al. \cite{Transformer} with a new design thought to improve it and adapt it as an element of a computer vision pipeline. We call the final architecture \textit{LHC-Net}: Local (multi-)Head Channel (self-attention) Network.
In the context of a wider research focused on the recognition of human emotions we tested \textit{LHC-Net} on the FER2013 dataset, a dataset for facial emotion recognition \cite{FER2013}. FER2013 was the object of a 2013 Kaggle competition. It is a dataset composed of $35587$ grey-scale $48x48$ images of faces classified in $7$ categories: anger, disgust, fear, happiness, sadness, surprise, neutral. The dataset is divided in a training set ($28709$ images), a public test set ($3589$ images), which we used as validation set, and a private test set ($3589$ images), usually considered the test set for final evaluations. FER2013 is known as a challenging dataset because of its noisy data with a relatively large number of non-face images and misclassifications. It is also strongly unbalanced, with only $436$ samples in the less populated category, ``Disgust'', and $7215$ samples in the more populated category, ``Happiness'':

\noindent\includegraphics[]{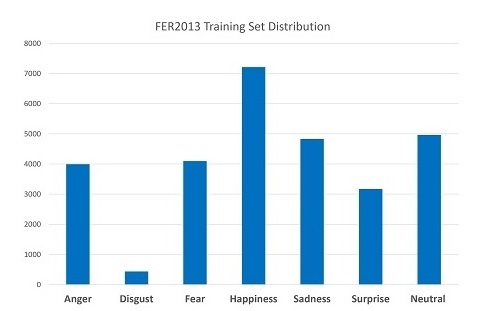}

\begin{figure*}[ht]\centering 
\includegraphics[width=\linewidth]{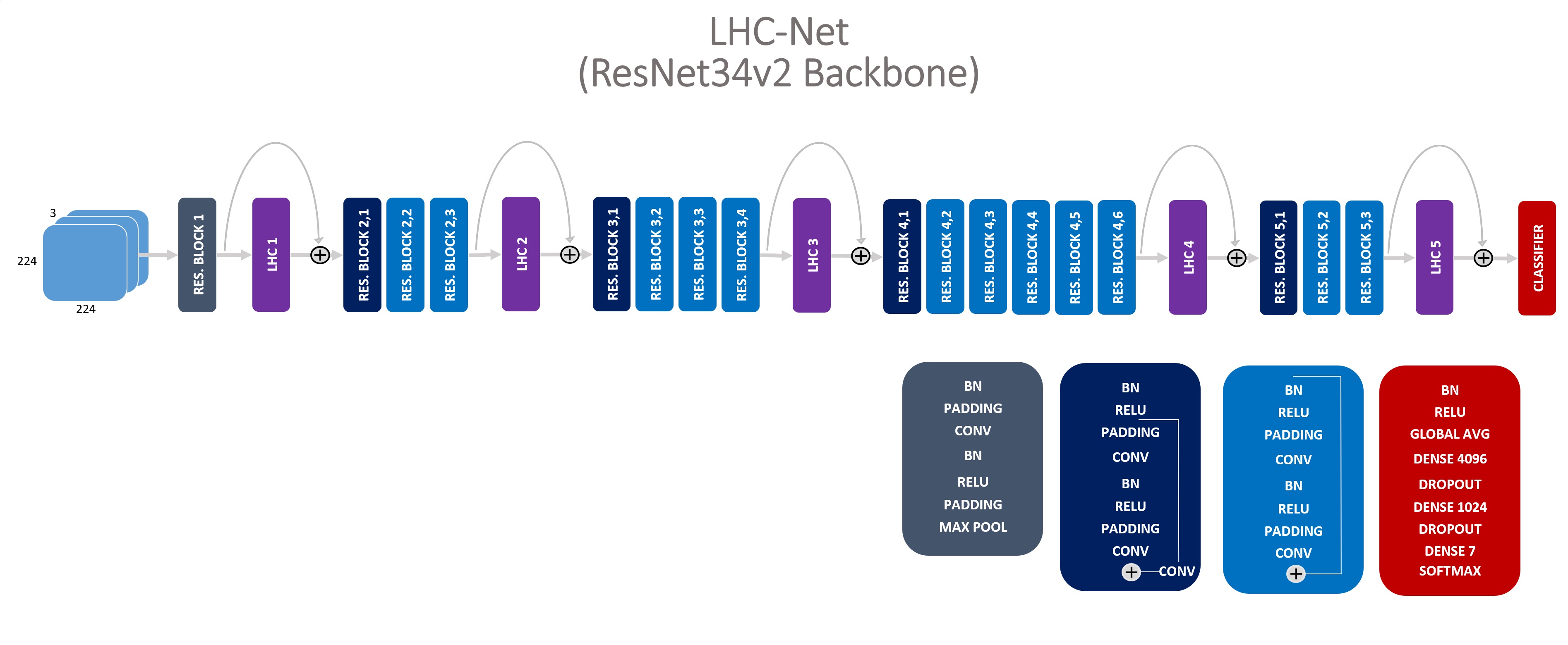}
\caption{Five \textit{LHC} modules integrated into a \textit{ResNet34v2} architecture. Every module features a residual connection to obtain an easier integration, especially when pre-training is used for the backbone architecture.}
\end{figure*}

\textit{LHC-Net} should be generally considered as a family of neural network architectures having a backbone represented by a convolutional neural network in which one or more \textit{LHC} modules are integrated. More specifically, in this paper, we will refer to \textit{LHC-Net} as a \textit{ResNet34} \cite{resnet} having integrated $5$ \textit{LHC} modules as shown in Figure 1.

For this reason \textit{LHC-Net} is a general purpose computer vision architecture since it doesn't feature any specific solution designed for facial expression recognition. 

In our experiments, \textit{LHC-Net} achieved a classification accuracy on the private test set of FER2013 which is, to the best of our knowledge (and accordingly with the paperswithcode's \href{https://paperswithcode.com/sota/facial-expression-recognition-on-fer2013}{\texttt{leaderboard}} at the time this paper is being written), the current single deep learning model state of the art both with and without test time augmentation, with a computational cost which is only a fraction of the previous \textit{SOTA} architecture.\textit{}


\section{Related work}
\subsection{Attention}
The attention paradigm became very popular in the last few years with a large variety of mechanics and implementations in both NLP and computer vision scenarios. There are two main attention paradigms: either we pay attention to the data with the idea of enhancing the most meaningful aspects or we can try to exploit the inner relationships within these aspects in order to produce a more meaningful representation of the data. The latter approach is usually called \textit{self-attention} because in some way the data pays attention to itself. 

The first approach was introduced in $2014$ by Bahdanau et al. \cite{bahdanau2014neural} an updated by Luong et al. in 2015 \cite{luong2015effective}. The proposed solutions were used in neural machine translation and integrated in a classic ``seq to seq'' encoder/decoder architecture in which the decoder learns what are the outputs of the encoder where to pay more attention dinamically. 
Self-attention was introduced in $2017$ by Vaswani et al. \cite{Transformer} (again for neural machine translation) and it is considered by many the greatest breakthrough technology in AI since the backpropagation was introduced in $1986$ \cite{rumelhart1986learning}. It fully replaced the previous state of the art technologies, the recurrent and convolutional networks, in NLP.

Since then there has been many attempts to bring self-attention in computer vision but, as of now, with only partial success. As opposite as the NLP case, in computer vision self-attention struggles to outperform the SOTA computer vision architectures like the classical \textit{Inception} \cite{inception},  \textit{ResNet} \cite{resnet}, \textit{VGG}, \cite{vgg} etc.

In computer vision there are several type of attentions paradigms, for clarity from now on we will use the following nomenclature:

\begin{itemize}
  \item \textbf{Global Attention}: usually it is only a module used before another model with the idea to enhance the important parts of an image and to ignore the rest of the image.
  \item \textbf{Spatial Attention}: the attention modules focus on single pixels or areas of the feature maps.
  \item \textbf{Channel Attention}: the attention modules focus on entire feature maps.
  \item \textbf{Self-Attention}: the attention tries to find relationships between different aspects of the data.
  \item \textbf{Stand-Alone Attention}: the architecture is aimed at fully replacing the convolutional blocks and defining a new processing block for computer vision based on some attention mechanism (mostly self-attention).
\end{itemize}

Xu et al. proposed a global attention module for medical image classification \cite{xu2020novel}, this module pre-processes images enhancing important areas pixel by pixel before feeding them into a standard convolutional neural network. This kind of pre-processing is thought to make more robust the following convolution processing. It could be associated to the one proposed by Jaderberg et al. \cite{jaderberg2015spatial} which attempts to compensate for the lack of rotation/scaling invariance of the convolution paradigm. The proposed module learns a sample-dependant affine transformation to be applied to images in order to make them centered and properly scaled/rotated. 

The channel approach we propose in this paper, despite being relatively unexplored in our self-attention mode, is instead very popular when associated with vanilla attention. Hu et al. proposed the \textit{SE-Net} (Squeeze and Excitation) \cite{hu2018squeeze}, a simple and effective module which enhances the more important features of a convolutional block. Squeeze and excitation lately became a key module in the very popular \textit{Efficient-Net} by Tan et al. \cite{tan2019efficientnet}  which set a new SOTA on several benchamrk datasets. 
Similarly Woo et al. proposed the \textit{CBAM} (Convolutional Block Attention Module), a sequential module composed of a spatial and a channel attention sub-modules \cite{woo2018cbam}.
There are other examples of channel and spatial vanilla attention: \textit{ECA-Net} (Efficient Channel Attention) by Wang et al. \cite{wang2020ecanet} is a new version of Squeeze and Excitation; \textit{SCA-CNN} (Spatial and Channel-wise attention) proposed by Chen et al. \cite{chen2017sca} combines both spatial and channel vanilla attention for image captioning. \textit{URCA-GAN} by Nie et al. \cite{nie2021urca} is a GAN (Generative Adversarial Network) featuring a residual channel attention mechanism thought for image-to-image translation.

Channel attention wasn't used only in vanilla approaches; similarly to our architecture Fu et al., Liu et al. and Tian et al. proposed self-attention architectures \cite{fu2019dual}, \cite{liu2021scsa}, \cite{tian2020triple} respectively for scene segmentation, feature matching between pairwise images and video segmentation. The main differences between these modules and ours are the following:

\begin{itemize}
  \item in all of them channel attention always has a secondary role and there's always a spatial attention sub-module with a primary role
  \item in all of them the crucial multi-head structure is lacking
  \item all of them implement channel attention as a ``passive'' non-learning module
  \item none of them integrates our local spatial behavior for channel attention
  \item none of them integrates our dynamic scaling which is very specific of our architecture.
\end{itemize}

As opposite as channel self-attention, spatial self-attention is widely explored, in most cases with the ambitious goal of totally replacing the convolution in computer vision, just like Vaswani's Transformer made \textit{LSTM} obsolete. 
Bello et al. proposed an attention-augmented convolutional network \cite{bello2019attention} in which Vaswani's self-attention is straightforwardly applied to pixels representations and integrated in a convolutional neural network.

Similarly Wu et al. proposed the Visual Transformer \cite{wu2020visual}, an architecture in which many ``tokens'' (i.e., image sections derived from a spatial attention module) are feeded into a transformer. The entire block is integrated in a convolutional network.
The Visual Transformer is inspired by \textit{ViT}, the Vision Transformer by Dosovitskiy et al. \cite{dosovitskiy2020image}, \textit{ViT} is a stand-alone spatial self-attention architecture in which the transformer's inputs are patches extracted from the tensor image. Previous attempts to implement stand-alone spatial attention were done by Ramachandran et al. \cite{ramachandran2019stand} and Parmar et al. \cite{parmar2018image}. Spatial self-attention was also used in GANs by Zhang et al. with their \textit{SAGAN} (Self-Attention Generative Adversarial Network) \cite{zhang2019self}.

More recently Liu et al. and Dai et al. proposed other two spatial stand-alone self-attention architectures, respectively the Swin Transformer \cite{liu2021swin} and the \textit{CoAtNet} \cite{dai2021coatnet} (depthwise Convolution and self-Attention). We can think at stand-alone architectures as attempts of rethinking convolution and replace it in a way able to address its limitations. Many improvements of convolution were proposed, mainly to make them invariant for more general transformations than translations, such as the Deep Simmetry Network proposed by Gens et al. \cite{gens2014deep} or the Deformable Convolutional Network by Dai et al. \cite{dai2017deformable}.

Both \textit{ViT} and \textit{CoAtNet} can be considered the current state of the art on Imagenet but they outperform Efficient Net by only a very small margin \cite{pham2021meta} and at the price of a complexity up to $30x$ and of a pre-training on the prohibitive JFT-3B dataset containing 3 billions of images.

These are good reasons for considering convolution not yet fully replaceable by self-attention in computer vision. But the main reason we didn't pursue the goal of a stand-alone architecture is that we don't believe in the main assumption spatial self-attention is based on in computer vision. Self-attention had a great success in NLP because it eventually exploited the inner relationships between the words in a phrase which sequential approaches were not able to model effectively. Every word in a sentence has a strong well defined relationship with any other word in that phrase, and they finally form a complex structure composed of these relationships. But, for instance, if we take a picture of a landscape we see no reason to believe that such a relationship could exist between a rock on the ground and a cloud in the sky or, even more extremely, between two random pixels, at least not in the same way the subject of a phrase is related to its verb.
On the other hand this observation does not hold for the features extracted from a picture, and the best way we know, so far, to extract features from a picture is convolution. These are the main reasons we decided to further explore channel self-attention in synergy with convolution, not as a stand-alone solution.


\subsection{FER2013}
As mentioned before FER2013 is a challenging dataset for facial expressions recognition. 
As reported by Goodfellow et al. even human accuracy on FER2013 is limited to $65\pm5\%$ \cite{FER2013}. Tang et al. \cite{FER2013} successfully used linear support vector machines reaching $71.16\%$ accuracy.
Minaee et al. achieved $70.02\%$ accuracy using a convolutional neural network augmented with a global spatial attention module \cite{minaee2021deep}. Pramerdorfer et al. experimented several architectures on FER2013 reaching $71.6\%$ accuracy with \textit{Inception}, $72.4\%$ with \textit{ResNet} and $72.7\%$ with \textit{VGG} \cite{pramerdorfer2016facial}. Khanzada et al. managed to achieve $72.7\%$ accuracy with \textit{SE-ResNet50} and $73.2\%$ with \textit{ResNet50} \cite{khanzada2020facial}. Khaireddin et al. reached $73.28\%$ accuracy using \textit{VGG} with a specific hyper-parameters fine tuning \cite{khaireddin2021facial}.
Pham et al. designed the \textit{ResMaskingNet} which is a \textit{ResNet} backbone augmented with a spatial attention module based on the popular \textit{U-Net}, a segmentation network mostly used in medical image processing. \textit{ResMaskingNet} achieves the remarkable accuracy of $74.14\%$. Pham et al. also reported that an ensemble of $6$ convolutional neural networks, including \textit{ResMaskingNet}, reaches $76.82\%$ accuracy \cite{resmaskingnet}.

\section{\textit{LHC-Net}}
As already mentioned and shown in Figure $1$ our \textit{LHC} module can be integrated in virtually any existing convolutional architecture, including of course \textit{AlexNet} \cite{krizhevsky2012imagenet}, \textit{VGG} \cite{vgg}, \textit{Inception} \cite{inception} and \textit{ResNet} \cite{resnet}. 

In this section we will give a detailed mathematical definition of \textit{LHC} as shown in Figure $2$, starting from a generic tensor and forward propagating it through the entire architecture.
\subsection{Architecture}
We first need to define the model hyper-parameters: let $n \in \mathbb{N}^{+}$ be the number of local heads, $s \in \mathbb{N}^{+}$ the kernel size of the convolution we will use to process the value tensor, $p \in \mathbb{N}^{+}$ the pool size used in average pooling and max pooling blocks, $d \in \mathbb{N}^{+}$ the embedding dimension of every head and $g \in \mathbb{R}^{\geq 0}$ a constant we will need in the dynamic scaling module.

Let $\mathbf{x} \in \mathbb{R}^{H,W,C}$ be a generic input tensor, where $H$, $W$ and $C$ are respectively the height, width and number of channels, with the constraint that $HxW$ must be divisible by $n$.

We define $\mathbf{Q}$, $\mathbf{K}$ and $\mathbf{V}$ as follows:

\begin{align}
&\mathbf{Q} = \operatorname{AvgPool}_{p,1}(\mathbf{x}) \in \mathbb{R}^{H,W,C}\\
&\mathbf{K} = \operatorname{MaxPool}_{p,1}(\mathbf{x}) \in \mathbb{R}^{H,W,C}\\
&\mathbf{V} = \operatorname{AvgPool}_{3,1}(\operatorname{2D-Conv}_{s,1}(\mathbf{x})) \in \mathbb{R}^{H,W,C}
\end{align}

\begin{figure*}[ht]\centering 
\includegraphics[width=\linewidth]{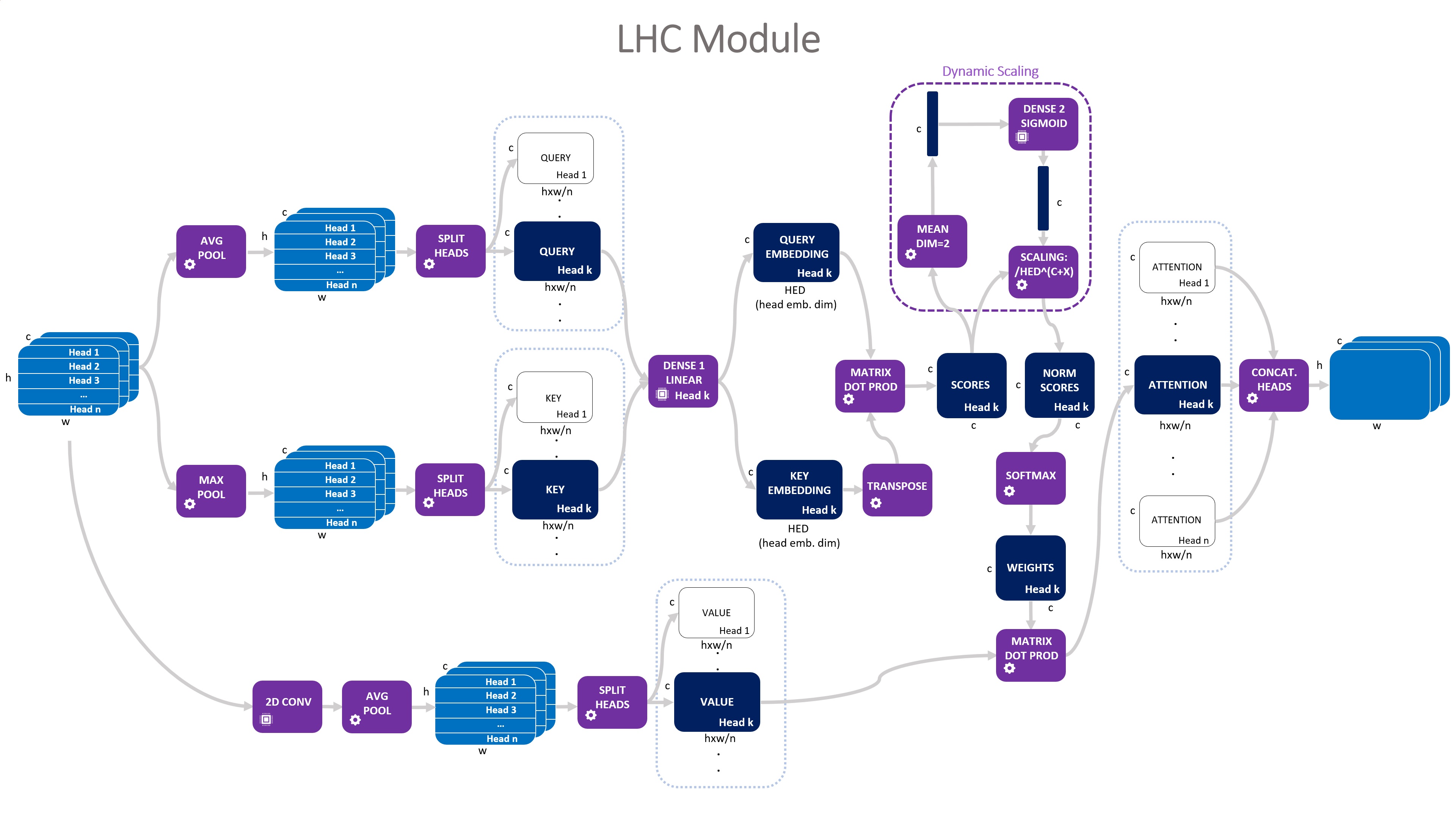}
\caption{The \textit{LHC} module in its more general multi-head form. Image tensors of shape $HxWxC$ are in pale blue, when reshaped/processed they are in dark blue. The processing units are in violet.}
\end{figure*}

where the pooling operators subscripts are respectively the pool size and the stride and the convolution operator subscripts are respectively the kernel size and the stride.

Now we want to split the tensors $\mathbf{Q}$, $\mathbf{K}$ and $\mathbf{V}$ into $n$ horizontal slices and reshape the resulting tensors as follows:
$\forall h = 1, ..., n$
\begin{align}
             &\mathbf{q}_{h} = [\operatorname{SplitHeads(\mathbf{Q})}]_h \in \mathbb{R}^{C,(HxW)/n}\\
             &\mathbf{k}_{h} = [\operatorname{SplitHeads(\mathbf{K})}]_h \in \mathbb{R}^{C,(HxW)/n}\\
             &\mathbf{v}_{h} = [\operatorname{SplitHeads(\mathbf{V})}]_h \in \mathbb{R}^{C,(HxW)/n}
\end{align}
Every head is deputed to process a triplet $(\mathbf{q}_{h}, \mathbf{k}_{h}, \mathbf{v}_{h})$ then we have $n$ separate fully connected layers with linear output and weights/biases: 
$\mathbf{w}_{1,h} \in \mathbb{R}^{(HxW)/n, d}$, $\mathbf{b}_{1,h} \in \mathbb{R}^{d}$.

Queries and keys will share the same dense blocks resulting in $n$ embeddings as follows:
\begin{align}
             & \hspace{1cm} \tilde{q}_{h}^{i,j} = \sum_{t=1}^{(HxW)/n}q_{h}^{i,t}w_{1,h}^{t,j} + b_{1,h}^{j} \in \mathbb{R}\\
             & \hspace{1cm} \tilde{k}_{h}^{i,j} = \sum_{t=1}^{(HxW)/n}k_{h}^{i,t}w_{1,h}^{t,j} + b_{1,h}^{j} \in \mathbb{R}\\
             & \hspace{4cm} \forall h = 1, ..., n \nonumber\\
             & \hspace{4cm} \forall i \, \,   = 1, ..., C \nonumber\\
             & \hspace{4cm} \forall j = 1, ..., d \nonumber
\end{align}

Or, more shortly (from now on we will omit the head logic quantifier):
\begin{align}
             & \hspace{1cm} \tilde{\mathbf{q}}_{h} = \mathbf{q}_{h} \cdot \mathbf{w}_{1,h} + \mathbf{b}_{1,h} \in \mathbb{R}^{C,d}\\
             & \hspace{1cm} \tilde{\mathbf{k}}_{h} = \mathbf{k}_{h} \cdot \mathbf{w}_{1,h} + \mathbf{b}_{1,h} \in \mathbb{R}^{C,d}
\end{align}

Now we can compute the attention scores through usual transposition and matrix product:

\begin{align}
             & \hspace{2cm} \mathbf{S}_{h} = \tilde{\mathbf{q}}_{h} \cdot  \tilde{\mathbf{k}}^{T}_{h} \in \mathbb{R}^{C,C}
\end{align}

Dynamic scaling produces a channel-wise learned scaling (not dependent from heads) through averaging the scores and passing them through another fully connected layer with sigmoid activation and weights/biases $\mathbf{w}_{2} \in \mathbb{R}^{C,C}$, $\mathbf{b}_{2} \in \mathbb{R}^{C}$:

\begin{align}
		   & \tilde{\mathbf{S}}_{h} = \operatorname{Mean}_{dim=2}(\mathbf{S}_{h}) \in \mathbb{R}^{C}\\[12pt]
             & T_{h}^{i} = \operatorname{Sig}\left(\sum_{t=1}^{C}\tilde{S}_{h}^{t}w_{2}^{t,i} + b_{2}^{i}\right) \in \mathbb{R} \, \, \, \, \, \,  \forall i =1, ..., C\\[2pt]
             & N_{h}^{i,j} = \frac{S_{h}^{i,j}}{d^{(g + T_{h}^{i})}} \in \mathbb{R} \, \, \, \, \, \,  \, \, \, \, \, \forall i, j =1, ..., C\\[12pt]
             & \mathbf{W}_{h} = \operatorname{Softmax}_{dim=2}(\mathbf{N}_{h}) \in \mathbb{R}^{C,C}
\end{align}
where $\mathbf{T}_{h}$ is the tensor of the scaling factors, $\mathbf{N}_{h}$ the tensor of the normalized attention scores and $\mathbf{W}_{h}$ the final attention weights associated with the head $h$.

Now we can compute the final attention tensor for head $h$ very straightforwardly:
\begin{align}
             & \hspace{2cm} \mathbf{A}_{h} = \mathbf{W}_{h} \cdot  \mathbf{v}_{h} \in \mathbb{R}^{C,(HxW)/n}
\end{align}
and using simple transpose, reshape and concatenation operators we can compose the output $\mathbf{y}$ by assembling the $n$ heads:
\begin{align}
             & \mathbf{y} = \operatorname{SplitHeads}^{-1}([\mathbf{A}_{1}, \mathbf{A}_{2}, ..., \mathbf{A}_{n}]) \in \mathbb{R}^{H,W,C}
\end{align}

\subsection{Motivation and Analysis}

\

\,

\paragraph{Channel Self-Attention}

\

We already explained the main reasons behind our choice of channel-wise self-attention. We can summarize them as follows:
\begin{itemize}
  \item spatial attention in computer vision strongly relies on the main assumption that a relationship between single pixels or areas of an image exists. This assumption is not self-evident or at least not as evident as the relationship between words in a phrase spatial attention is inspired by
  \item all attempts to pursue spatial self-attention in computer vision (especially in stand-alone mode) gained only minor improvements over previous state of the art architectures and, most of the times, at the price of an unreasonably higher computational cost and a prohibitive pre-training on enormous datasets
  \item much more simple and computationally cheaper approaches, like Squeeze and Excitation in \textit{Efficient Net}, are already proven to be very effective without the need to replace convolution  
\end{itemize}

\

\paragraph{Dynamic Scaling}

\

In Vaswani's \textit{Transformer} the scaling is static and constant among the sequence. Equation $(14)$ becomes:
$$\mathbf{N} = \frac{\mathbf{S}}{\sqrt{d}}$$
The idea behind our dynamic scaling is to exploit the following behavior of $\operatorname{Softmax}$ function. Given a non constant vector $\mathbf{x} \in \mathbb{R}^n$ and a positive constant $\alpha > 0$ it results:
\begin{align}
		   & \lim_{\alpha \to +\infty}\frac{\displaystyle \mathrm{e}^{\alpha x_{i}}}{\displaystyle \sum_{j=1}^{n}\mathrm{e}^{\alpha x_{j}}} = \begin{cases} 1, & \mathrm{if} \hspace{0.1cm} x_{i} = \operatorname{max}(\mathbf{x}) \\ 0, & \mathrm{otherwise} \end{cases} \\
             & \lim_{\alpha \to 0^{+}}\frac{\displaystyle \mathrm{e}^{\alpha x_{i}}}{\displaystyle \sum_{j=1}^{n}\mathrm{e}^{\alpha x_{j}}} = \frac{\displaystyle 1}{\displaystyle \sum_{j=1}^{n}1} = \frac{1}{n}\\
             & \frac{\displaystyle \mathrm{e}^{x_{i_1}}}{\displaystyle \sum_{j=1}^{n}\mathrm{e}^{x_{j}}} < \frac{\displaystyle \mathrm{e}^{x_{i_2}}}{\displaystyle \sum_{j=1}^{n}\mathrm{e}^{x_{j}}} \Leftrightarrow 
                \frac{\displaystyle \mathrm{e}^{\alpha x_{i_1}}}{\displaystyle \sum_{j=1}^{n}\mathrm{e}^{\alpha x_{j}}} < \frac{\displaystyle \mathrm{e}^{\alpha x_{i_2}}}{\displaystyle \sum_{j=1}^{n}\mathrm{e}^{\alpha x_{j}}}
\end{align}

These equations imply that we can multiply a logits vector $\mathbf{x}$ by a positive constant $\alpha$ without altering its softmax ranking (equation $(20)$) and if $\alpha$ is small the softmax resulting vector approximates an arithmetic average (equation $(19)$), if it is large it will be close to a one-hot vector valued $1$ on the max of $\mathbf{x}$ and $0$ otherwise (equation $(18)$).
In other words the dynamic scaling module learns how complex the new feature maps must be. If the $\alpha$ associated to a given new feature map is large this feature map will be a strict selection of old feature maps, if it is small the new feature map will be a more complex composition of old feature maps involving a greater number of them.

\

\paragraph{Shared Linear Embedding and Convolution}

\

A shared linear layer was already explored by Woo et al. with their \textit{CBAM} vanilla attention architecture \cite{woo2018cbam}. Our idea is exploiting the ``self'' nature of our attention mechanism. Using Vaswani's terminology self-attention means that the query, key and value originate from the same tensor. We decided to leverage this aspect and save some complexity by first differentiating query and key respectively with average and max pooling in order to enhance different scale aspects of the input and then feeding them into a single shared linear embedding layer. Dense mapping is also helped by the big dimensionality reduction due to head splitting.

On the other hand we used global convolution for the entire value tensor in order to preserve the bi-dimensional structure of the data. 

\

\paragraph{Local Multi-Head}

\

In the original \textit{Transformer} the multi-head structure is a concatenation of blocks all processing the same input. Voita et al. \cite{voita2019analyzing} analyzed the \textit{Transformer} and found a surprising redundancy in the representation offered by different heads: pruning $44$ out of $48$ heads from the \textit{Transformer}'s encoder block results only in a residual performance drop. Only $4$ heads ($8\%$) were necessary to maintain a performance very close to the one of the entire architecture. We tried to perform a similar evaluation for the \textit{LHC-Net} by simply ``switching off'' \textit{LHC} blocks and by ``de-training'' them (i.e., set the weights and biases of \textit{LHC} blocks at the initialization status, before training). In our case it was feasible to just switch off or de-train the new blocks without any further training because the entire \textit{ResNet} backbone of the network was pre-trained and already able to deliver a very high performance. With this approach we found that at least $16$ heads out of $31$ ($52\%$) were necessary, more precisely the first $2$ \textit{LHC} blocks.

We further analyzed this behavior and in order to make another comparison we trained a standard \textit{Transformer} encoder block as a simple classifier for a NLP classification problem reaching a very high accuracy, then we evaluated the model by simple correlation between the output of the heads and found a correlation between heads up to $93\%$. As a comparison our architecture had a correlation between heads of $63\%$.

\

There are many attempts to improve the attention mechanism of the \textit{Transformer}. Cordonnier et al. tried to address the redundancy issue with a modified multi-head mechanism \cite{cordonnier2020multi}. Sukhbaatar et al. proposed a modified \textit{Transformer} with an adaptive attention span \cite{sukhbaatar2019adaptive}.

More similarly to our local approach India et al. proposed a multi-head attention mechanism for speaker recognition in which every head processes a different section of the input \cite{india2019self}. There are two main differences with our approach (other than application field and implementation details): 
\begin{itemize}
 \item Their approach is not designed for self-attention
 \item Their local processing units are used at a later stage. They directly calculate local attention weights from embeddings (scalar output with softmax activation). Our local processing units calculate the initial embeddings (high dimension output with linear activation)
\end{itemize}

\

The ideas behind local heads are mainly three:
\begin{itemize}
 \item Local heads have the advantage of working at a much lower dimension. Detecting a pattern of few pixels is harder if the input includes the entire feature map
 \item Splitting the images in smaller parts gives to local heads the ability to build new feature maps considering only the important parts of the old maps. There's no reason to compose feature maps in their entirety when only a small part is detecting an interesting feature. Local heads are able to add a feature map to a new feature map only if the original map is activated by a pattern and only around that pattern, avoiding then to add not useful informations
 \item Local heads seem to be more efficient in terms of parameters allocation
\end{itemize}

We experimentally found the performance positively correlated with the number of heads but we also tried to give a qualitative explanation of the third observation by designing a concrete example. Let's say we have $n$ feature maps as output of the previous convolution block and that the optimal composition of those maps includes a combination of $2$ of them, the  $i^{th}$ and the $j^{th}$, in the $k^{th}$ target feature map. In order to learn this pattern, using equations $(9)$ and $(10)$ (omitting the biases), a single global head must map:

\begin{align}
             & \hspace{1.55cm} \tilde{\mathbf{q}} = \mathbf{q} \cdot \mathbf{w}_{1} \in \mathbb{R}^{C,d} \nonumber \\
             & \hspace{1.55cm} \tilde{\mathbf{k}} = \mathbf{k} \cdot \mathbf{w}_{1} \in \mathbb{R}^{C,d} \nonumber
\end{align}

in such a way that $\tilde{\mathbf{q}}_k$ and $\tilde{\mathbf{k}}_i$ must be collinear in order to produce a high attention score in the $k^{th}$ target feature for the $i^{th}$ old feature map by dot product. The same for $\tilde{\mathbf{q}}_k$ and $\tilde{\mathbf{k}}_j$.
To summarize we have $3$ vectors that need to be mapped in other $3$ vectors linked by $2$ constraint rules. In total we have $3(HW + d)$ dimensions or $3(HWd)$ relationships subject to $2$ constraints  to be modeled. To do that with the embedding linear layer we have a matrix $\mathbf{w} \in \mathbb{R}^{HxW,d}$, equivalent to $HWd$ free parameters. So we have: 
\begin{align}
             & \hspace{1.55cm} G1 = \frac{HWd}{3(HW + d)2} \\
             & \hspace{1.55cm} G2 = \frac{HWd}{3(HWd)2} = \frac{1}{6}
\end{align}

where $G1$ is the number of free parameters for dimension for every constraint and $G2$ is the number of free parameters for relationship for every constraint. We see them as qualitative measures of the efficiency of the global single head.
Now we want to calculate them in the case of $n$ local heads. The difference is that local heads works only on fractions of the entire input tensor, so we have to take into account where the $i^{th}$ and the $j^{th}$ filters are eventually activated. For a given section of the input tensor there are $3$ cases: only one of them could be activated in that area, both of them or none of them. We call $A$ the number of sections with $1$ possible activation, $B$ the number of sections with $2$ possible activations and $C$ the number of sections with no possible activations. It results:

$$ A + B + C = n $$

but this time $\mathbf{w}_{1,h} \in \mathbb{R}^{(HxW)/n, d}$, hence we have:

\begin{align}
             L1 &= \left(A\frac{\frac{HW}{n}d}{2(\frac{HW}{n} + d)} +  B\frac{\frac{HW}{n}d}{3(\frac{HW}{n} + d)2}\right)/(A + B)\\
             L2 &= \left(A\frac{\frac{HW}{n}d}{2(\frac{HW}{n}d)} +  B\frac{\frac{HW}{n}d}{3(\frac{HW}{n}d)2}\right)/(A + B) \nonumber \\
                 &= \left(\frac{A}{2} +  \frac{B}{6}\right)/(A + B)
\end{align}

We have immediately:
\begin{align}
             & L2 > G2 \Leftrightarrow \nonumber\\
             & \left(\frac{A}{2} +  \frac{B}{6}\right)/(A + B) > \frac{1}{6} \Leftrightarrow A > 0 \nonumber                 
\end{align}

Or more shortly: 
\begin{align}
             & \hspace{1.55cm} L2 \geq G2\\
             & \hspace{1.55cm} L2 = G2 \Leftrightarrow A = 0                
\end{align}

if the $i^{th}$ and the $j^{th}$ filters are possibly activated in every section of the input tensor local multi-head is equivalent to global single head in terms of efficiency and effectiveness, but a single section of the input tensor with only one possible activation is enough to make local multi-head more effective.

If we decide to consider the dimensions ($L1$ and $G1$ measures instead of $L2$ and $G2$) the calculation is more complicated; to make it easier let's make some basic assumptions. Let's consider the hyper-parameters settings of the actual first two blocks of our \textit{LHC-Net}, where $d = \frac{HW}{2n}$ and $n = 8$. We have:

\begin{align}
             &L1 > G1 \Leftrightarrow \nonumber \\
             &\frac{\left(A\frac{\frac{HW}{n}d}{2(\frac{HW}{n} + d)} +  B\frac{\frac{HW}{n}d}{2(\frac{HW}{n} + d)3}\right)}{(A + B)}  > \frac{HWd}{3(HW + d)2} \Leftrightarrow \nonumber\\
             &\left(A\frac{\frac{1}{n}}{2\frac{HW}{n}(1 + \frac{1}{2})} +  B\frac{\frac{1}{n}}{6\frac{HW}{n}(1 + \frac{1}{2})}\right)  > \frac{A+B}{6HW(1 + \frac{1}{2n})} \Leftrightarrow \nonumber \\
             &\left(A\frac{1}{2(1 + \frac{1}{2})} +  B\frac{1}{6(1 + \frac{1}{2})}\right)  > \frac{A+B}{6(1 + \frac{1}{16})} \Leftrightarrow \nonumber \\
             &\left(\frac{A}{3} + \frac{B}{9}\right)  > \frac{16A+16B}{102} \Leftrightarrow \nonumber \\
             &\left(3A + B\right)  > \frac{144A+144B}{102} \Leftrightarrow A > 0.26B\nonumber
\end{align}

In this case the combinations $A=0$, $B=8$ and $A=1$, $B=7$ give an advantage to global single head. Every other possible combination do the opposite as shown in this figure:

\noindent\includegraphics[]{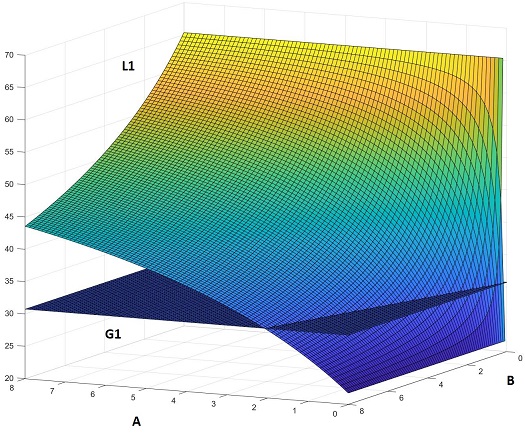}

It appears clear that local heads have an advantage over global heads in any real-world application. For example in FER2013 it is unlikely that a feature extracted from a face could appear anywhere in the picture. For example eyebrows will be almost always in the upper section of the picture.

This, of course, has not the ambition to be a rigorous proof of the goodness of local heads over global head, it is only a qualitative analysis giving an encouraging view.

\section{Experiments}

As mentioned we mainly focused on using \textit{LHC} in conjunction with a pre-trained backbone, the \textit{ResNet34v2}. The training process consisted in training a \textit{ResNet34v2} (with Imagenet pre-training initialization) on FER2013, then adding $5$ \textit{LHC} modules as shown in Fig.1 and further training the entire architecture. The Idea was designing modules with a small impact on the ``host'' network similarly at the approach of the Squeeze and Excitation modules \cite{hu2018squeeze}. In other words our main goal was to test the ability of \textit{LHC} to give an extra performance boost to an existing good performing model. Secondarily we also tested \textit{LHC-Net} as a stand-alone model trained (only Imagenet pre-training of the \textit{ResNet} part) from scratch and we obtained limited but very good results. In this section we will discuss the details of our training protocol and the experimental results.
\

\paragraph{Setup}

\

We rescaled the FER2013 images to $224x224$ and converted them to RGB in order to match the resolution of Imagenet and make them compatible with the pre-trained \textit{ResNet}. For rescaling we used bilinear interpolation.
Then, in order to save RAM memory, we stored the entire training set as jpeg images accepting some neglectable quality loss and used TensorFlow Image Data Generator to feed the model during training. Saving images in jpeg format implies two different quality losses: the jpeg compression itself and the need to approximate the tensor images to be uint8 (bilinear interpolation in rescaling generate non integer values). To do that the tensors could be rounded or truncated. Considering that truncation is only a rounding with the input shifted of $0.5$ and that this shifting makes the training set in FER2013 better matching the validation and test set average pixel value we proceeded with raw truncation.

The implementation details of \textit{ResNet} are reported in Fig.1 and the model parameters of the $5$ \textit{LHC} blocks are the following:
\begin{center}
\begin{tabular}{||c | c | c | c | c | c||}  
 \hline
 \textbf{Block}  & \textbf{Heads} & \textbf{Dim} & \textbf{Pool} & \textbf{Scale} & \textbf{Ker}\\ 
 \hline
 \hline
\textbf{\textit{LHC}$1$} & $8$ & $196$ & $3$ &$1$ &$3$ \\
 \hline
 \hline
\textbf{\textit{LHC}$2$} & $8$ & $196$ & $3$ &$1$ &$3$ \\
 \hline
 \hline
\textbf{\textit{LHC}$3$} & $7$ & $56$ & $3$ &$1$ &$3$ \\
 \hline
 \hline
\textbf{\textit{LHC}$4$} & $7$ & $14$ & $3$ &$1$ &$3$ \\
 \hline
 \hline
\textbf{\textit{LHC}$5$} & $1$ & $25$ & $3$ &$1$ &$3$ \\
 \hline
\end{tabular}
\end{center}

We trained the model in a sequential mode with $3$ training stages, using standard crossentropy loss, varying the data augmentation protocol, the batch size and the optimzer at every stage. Early stopping is performed on validation set.

\

\textbf{Stage1}:
\begin{center}
\begin{tabular}{||c c||}  
 \hline
 \textbf{Optimizer} & Adam, lr = $0.0001$ \\ 
 \hline
 \textbf{Batch Size} & $48$ \\
 \hline
 \textbf{Patience} & $30$ epochs \\
 \hline
 \textbf{Augmentation} & $30$ deg. rot. \\
 \hline
\end{tabular}
\end{center}

\textbf{Stage2}:
\begin{center}
\begin{tabular}{||c c||}  
 \hline
 \textbf{Optimizer} & SGD, lr = $0.01$ \\ 
 \hline
 \textbf{Batch Size} & $64$ \\
 \hline
 \textbf{Patience} & $10$ epochs \\
 \hline
                                 & $10$ deg. rot. \\
  \textbf{Augmentation} & $0.1$ h/v shift \\
                                 & $0.1$ zoom \\
 \hline
\end{tabular}
\end{center}

\textbf{Stage3}:

\begin{center}
\begin{tabular}{||c c||}  
 \hline
 \textbf{Optimizer} & SGD, lr = $0.01$ \\ 
 \hline
 \textbf{Batch Size} & $64$ \\
 \hline
 \textbf{Patience} & $5$ epochs \\
 \hline
 \textbf{Augmentation} & - \\
 \hline
\end{tabular}
\end{center}

\

At this point we have our \textit{ResNet} ready to be augmented and further trained. We used a very simple training protocol.

\

\textbf{Stage 4 (\textit{LHC} training)}:

\begin{center}
\begin{tabular}{||c c||}  
 \hline
 \textbf{Optimizer} & SGD, lr = $0.01$ \\ 
 \hline
 \textbf{Batch Size} & $64$ \\
 \hline
 \textbf{Patience} & $3$ epochs \\
 \hline
 \textbf{Augmentation} & - \\
 \hline
\end{tabular}
\end{center}
We observed in some cases, depending on the \textit{LHC} initialization, that the added modules are somehow ``rejected'' by the host network and the training struggles to converge, in one case it totally diverged.
It happened in a minority of the total attempts but to perform the following evaluations we kept only the models whose training loss was less than the starting \textit{ResNet} training loss plus an extra $10\%$ to take into account the augmented complexity of the model.

To evaluate \textit{LHC} we first applied stage $4$ to the single best \textit{ResNet34} model we managed to achieve (with stages $1$, $2$ and $3$), varying the data generator seed, without \textit{LHC} modules (set $A$). Then, starting from the same base network we augmented it with \textit{LCH} modules and trained it using the same protocol. We tried a small number of trainings with a variety of model parameters (keeping the data generator seed fixed) and clearly detected a neighbourhood of settings appearing to work well (set $B$). At this point we trained several other models with the best promising parameters setting varying the generator seed (set $C$). We then compared the set $A$ with the set $B \cup C$.

We also considered a minor variation of \textit{LHC-Net}. We tried to exploit the analysis on the $5$ modules we discussed in the previous section showing the last modules playing a minor role and trained $5$ weights, limited by hyperbolic tangent, for every residual sum shown in Fig.1. We manually initialized this $5$ weights by setting them as follows: $a_1 = \operatorname{tansig}(0)$, $a_2 = \operatorname{tansig}(0)$, $a_3 = \operatorname{tansig}(0)$, $a_4 = \operatorname{tansig}(-1)$, $a_5 = \operatorname{tansig}(-0.5)$ with the idea of limiting the impact of the last $2$ modules. We call it \textit{LHC-NetC}.
Accordingly with the original Kaggle rules and with almost all evaluation protocols in literature only the private test set was used for final evaluations (public test set performance also appeared to be not well correlated with neither training nor private test performances).
For comparison with \textit{ResNet} we didn't use test time augmentation (TTA). We used TTA only for final evaluation and comparison with other models in literature.
Our TTA protocol is totally deterministic; we first used a sequence of transformations involving horizontal flipping, $\pm 10$ pixels horizontal/vertical shifts and finally $\pm 0.4$ radians rotations, in this order. We use rotation after shifting to combine their effect. Rotating first puts the images in only 9 spots, which becomes $25$ if we shift first. At this point we used a second batch of transformations involving horizontal flipping, $10\%$ zoom and again $\pm 0.4$ radians rotations. Finally we weighted the no-transformation inference $3$ times the weight of others inferences.

\paragraph{Results}
\begin{center}
\begin{tabular}{||p{1.58cm} | p{0.85cm} | p{0.85cm} | p{0.85cm} | p{0.85cm} | p{0.85cm}||}  
 \hline
                      &                                                & \hspace{0.15cm}\textbf{Top}         &                                                    &\hspace{0.08cm} \textbf{Top}                             &  \\ 
                      &\hspace{0.1cm} \textbf{Top}        & \hspace{0.15cm}$\mathbf{40 \%}$ & \hspace{0.15cm}\textbf{Top}           &\hspace{0.08cm} $\mathbf{25 \%}$                    &\\
\hspace{0.35cm}\textbf{Model}                      &\hspace{0.1cm} $\mathbf{40 \%}$ & \hspace{0.15cm}\textbf{w/o}         & \hspace{0.15cm}$\mathbf{25 \%}$   &\hspace{0.08cm} \textbf{w/o}                            & \hspace{0.03cm} \textbf{Best}\\
                      &                                                & \hspace{0.15cm}\textbf{best}         &                                                    &\hspace{0.14cm}\textbf{best}                            &\\
 \hline
 \hline
\textit{ResNet34v2} & $72.69\%$ & $72.65\%$ & $72.75\%$ & $72.69\%$ & $72.81\%$\\
 \hline
 \hline
\textbf{\textit{LHC-Net}} & $\mathbf{72.89\%}$ & $72.77\%$ & $\mathbf{73.02\%}$ & $72.83\%$ & $\mathbf{73.39\%}$\\
 \hline
 \hline
\textbf{\textit{LHC-NetC}} & $\mathbf{73.04\%}$ & $72.79\%$ & $\mathbf{73.21\%}$ & $72.89\%$ & $\mathbf{73.53\%}$\\
 \hline
\end{tabular}
\end{center}

\textit{LHC-Net} was able to consistently outperform our best performing \textit{ResNet34v2},  both on average and on peak result. Note that the average is not dramatically affected by peak result. Removing peak results does not alter the average qualitative evaluation. 
\begin{center}
\begin{tabular}{||p{2.2cm} | p{1.3cm} | p{0.7cm} | p{1.2cm} | p{0.8cm}||}  
\hline
\textbf{Model} & \textbf{Accuracy} & \textbf{TTA} & \textbf{Params} & \hspace{0.17cm}\textbf{Att}\\ 
\hline
\hline
\textit{BoW Repr.}\cite{FER2013}          & \hspace{0.15cm}$67.48\%$               & \hspace{0.17cm}no  & \hspace{0.5cm}-           & \hspace{0.4cm}-\\
\hline
\hline
Human \cite{FER2013}                         & \hspace{0.4cm}70\%                      & \hspace{0.15cm}no  & \hspace{0.5cm}-           & \hspace{0.4cm}- \\
\hline
\hline
\textit{CNN}\cite{minaee2021deep}       & \hspace{0.15cm}$70.02\%$               & \hspace{0.15cm}no   & \hspace{0.5cm}-          & \hspace{0.4cm}-\\
\hline
\hline
\textit{VGG19}\cite{resmaskingnet}       & \hspace{0.15cm}$70.80\%$               & \hspace{0.12cm}yes   & \hspace{0.05cm}$143.7$M & \hspace{0.4cm}-\\
\hline
\hline
\textit{EffNet}\cite{resmaskingnet}$^{*}$       & \hspace{0.15cm}$70.80\%$               & \hspace{0.12cm}yes   & \hspace{0.12cm}$9.18$M &\hspace{0.4cm}-\\
\hline
\hline
\textit{SVM}\cite{FER2013}                  & \hspace{0.15cm}$71.16\%$               & \hspace{0.15cm}no   & \hspace{0.5cm}- & \hspace{0.4cm}-\\
\hline
\hline
\textit{Inception}\cite{pramerdorfer2016facial}       & \hspace{0.15cm}$71.60\%$               & \hspace{0.12cm}yes   & \hspace{0.05cm}$23.85$M & \hspace{0.4cm}-\\
\hline
\hline
\textit{Incep.v1}\cite{resmaskingnet}$^{*}$       & \hspace{0.15cm}$71.97\%$               & \hspace{0.12cm}yes   & \hspace{0.32cm}$5$M & \hspace{0.4cm}-\\
\hline
\hline
\textit{ResNet34}\cite{pramerdorfer2016facial}       & \hspace{0.15cm}$72.40\%$               & \hspace{0.12cm}yes   & \hspace{0.12cm}$27.6$M & \hspace{0.4cm}-\\
\hline
\hline
\textit{ResNet34}\cite{resmaskingnet}       & \hspace{0.15cm}$72.42\%$               & \hspace{0.12cm}yes   & \hspace{0.12cm}$27.6$M & \hspace{0.4cm}-\\
\hline
\hline
\textit{VGG}\cite{pramerdorfer2016facial}       & \hspace{0.15cm}$72.70\%$               & \hspace{0.12cm}yes   & \hspace{0.05cm}$143.7$M & \hspace{0.4cm}-\\
\hline
\hline
\textit{SE-Net50}\cite{khanzada2020facial}       & \hspace{0.15cm}$72.70\%$               & \hspace{0.12cm}yes   & \hspace{0.27cm}$27$M & $5.18\%$\\
\hline
\hline
\textit{Incep.v3}\cite{resmaskingnet}$^{*}$       & \hspace{0.15cm}$72.72\%$               & \hspace{0.12cm}yes   & \hspace{0.05cm}$23.85$M & \hspace{0.4cm}-\\
\hline
\hline
\textit{ResNet34v2}       & \hspace{0.15cm}$72.81\%$               & \hspace{0.15cm}no   & \hspace{0.12cm}27.6M & \hspace{0.4cm}-\\
\hline
\hline
\textit{BAMRN50}\cite{resmaskingnet}$^{*}$       & \hspace{0.15cm}$73.14\%$               & \hspace{0.12cm}yes   & \hspace{0.05cm}$24.07$M & $1.62\%$\\
\hline
\hline
\textit{Dense121}\cite{resmaskingnet}       & \hspace{0.15cm}$73.16\%$               & \hspace{0.12cm}yes   & \hspace{0.12cm}$8.06$M & \hspace{0.4cm}-\\
\hline
\hline
\textit{ResNet50}\cite{khanzada2020facial}       & \hspace{0.15cm}$73.20\%$               & \hspace{0.12cm}yes   & \hspace{0.12cm}$25.6$M & \hspace{0.4cm}-\\
\hline
\hline
\textit{ResNet152}\cite{resmaskingnet}       & \hspace{0.15cm}$73.22\%$               & \hspace{0.12cm}yes   & \hspace{0.05cm}$60.38$M & \hspace{0.4cm}-\\
\hline
\hline
\textit{VGG}\cite{khaireddin2021facial}       & \hspace{0.15cm}$73.28\%$               & \hspace{0.12cm}yes   & \hspace{0.05cm}$143.7$M & \hspace{0.4cm}-\\
\hline
\hline
\textit{CBAMRN50}\cite{resmaskingnet}       & \hspace{0.15cm}$73.39\%$               & \hspace{0.12cm}yes   & \hspace{0.05cm}$28.09$M & \hspace{0.2cm}$9\%$\\
\hline
\hline
\textbf{\textit{LHC-Net}}                      & \hspace{0.15cm}$\mathbf{73.39\%}$ & \hspace{0.15cm}no  & \hspace{0.12cm}$32.4$M & $14.8\%$\\
\hline
\hline
\textbf{\textit{LHC-NetC}}                    & \hspace{0.15cm}$\mathbf{73.53\%}$ & \hspace{0.15cm}no  & \hspace{0.12cm}$32.4$M & $14.8\%$\\
\hline
\hline
\textit{ResNet34v2}       & \hspace{0.15cm}$73.92\%$               & \hspace{0.12cm}yes   & \hspace{0.12cm}$27.6$M & \hspace{0.4cm}-\\
\hline
\hline
\textit{RM-Net} \cite{resmaskingnet}      & \hspace{0.15cm}$74.14\%$              & \hspace{0.12cm}yes & \hspace{0.05cm}142.9M & $80.7\%$ \\
\hline
\hline
\textbf{\textit{LHC-NetC}}                    & \hspace{0.15cm}$\mathbf{74.28\%}$ & \hspace{0.12cm}yes & \hspace{0.12cm}$32.4$M & $14.8\%$\\
\hline
\hline
\textbf{\textit{LHC-Net}}                      & \hspace{0.15cm}$\mathbf{74.42\%}$ & \hspace{0.12cm}yes & \hspace{0.12cm}$32.4$M & $14.8\%$\\
\hline
\end{tabular}
\end{center}
$^{*}$ these models are reported in the GitHub repository associated with the referenced paper, not directly into the paper.
\

There are some key points emerging from this analysis:
\begin{itemize}
\item \textit{ResNet34} is confirmed to be the most effective architecture on FER2013, especially its v$2$ version. In our experiments raw \textit{ResNet34} trained with the multi-stage protocol and inferenced with TTA reaches an accuracy not distant from the previous SOTA (\textit{ResMaskingNet})
\item heavy architectures seem not able to outperform more simple models on FER2013
\item \textit{LHC-Net} has the top accuracy both with and without TTA
\item \textit{LHC-NetC} outperforms \textit{LHC-Net} but is outperformed when TTA is used
\item more importantly, \textit{LHC-Net} outperforms the previous SOTA with less than one fourth of its free parameters and the impact of the \text{LHC} modules on the base architecture is much lower (less than $15\%$ VS over $80\%$) and it is close to other attention modules like \textit{CBAM}/\textit{BAM}/\textit{SE}
\end{itemize}

As mentioned we limitedly experimented stand-alone training as well with very good results. We trained in parallel, using the same data generator seeds and the same multi-stage protocol, both \textit{LHC-Net} and \textit{ResNet34v2}. In both models the \textit{ResNet34v2} was initialized with Imagenet pre-trained weights. It resulted that \textit{LHC-Net} consistently outperformed \textit{ResNet34v2} at the end of every training stage. It is a limited but very encouraging result.

\section{Conclusions and Future Developments}
Attention, in its every form and shape, is a powerful idea and, despite its impact on computer vision might be not as revolutionary as on NLP, it is still proven to be an important, sometimes decisive, tool.

\

In particular we designed a novel local multi-head channel self-attention module, the \textit{LHC},  and it contributed proving that channel self-attention, in synergy with convolution, could be a functioning paradigm by setting a new state of the art on the well known FER2013 dataset. We also proved that self-attention works well as a small attention module intended as a booster for pre-existing architectures, like other famous vanilla attention modules as \textit{CBAM} or Squeeze and Excitation.

The future research on this architectures will include many aspects:

\begin{itemize}
\item testing \textit{LHC} on other, more computational intensive, scenarios like the Imagenet dataset
\item testing \textit{LHC} with other backbone architectures and with a larger range of starting performances (not only peak performances)
\item we did not optimize the general topology of \textit{LHC-Net} and the model hyper-parameters of the attention blocks are hand-selected with only a few attempts. There's evidence that both the $5$ blocks topology and hyper-parameters might be sub-optimal
\item further research on the stand-alone training mode will be necessary
\item normalization blocks before and after the the \textit{LHC} blocks should be better evaluated in order to mitigate the divergence issue mentioned in the previous section
\item a second convolution before the residual connection should be considered to mimic the general structure of the original \textit{Transformer}
\item a better head splitting technique could be key in the future research. The horizontal splitting we used was only the most obvious way to do it but not necessarily the most effective. Other approaches should be evaluated. For example learning the optimal areas through spatial attention
\end{itemize}

The main results of this paper are replicable by cloning the repository and following the instructions available at:
\
\href{https://github.com/Bodhis4ttva/LHC_Net}{\texttt{https://github.com/Bodhis4ttva/LHC\textunderscore Net}}

\section{Acknowledgment}
We would like to express our deepest appreciation to Dr. Carmen Frasca for her crucial support to our research. We would also like to extend our sincere thanks to Dr. Luan Pham and Valerio Coderoni for their helpfulness and kindness.

\newpage

\

\newpage

\phantomsection
\bibliographystyle{unsrt}
\bibliography{references}


\end{document}